\documentclass[letterpaper]{article} 
\usepackage[draft]{aaai2026}  
\usepackage{newtxtext,newtxmath} 
\usepackage{helvet}  
\usepackage{courier}  
\usepackage[hyphens]{url}  
\usepackage{graphicx} 
\usepackage{natbib}  
\usepackage{caption} 
\usepackage{amsmath}

\usepackage{amssymb}
\usepackage{booktabs}
\usepackage{multirow}
\usepackage[table]{xcolor}
\definecolor{lightgray}{gray}{0.9}
\usepackage{algorithm}
\usepackage{algorithmic}

\usepackage{newfloat}
\usepackage{listings}
\DeclareCaptionStyle{ruled}{labelfont=normalfont,labelsep=colon,strut=off} 
\floatstyle{ruled}
\newfloat{listing}{tb}{lst}{}
\floatname{listing}{Listing}
\title{CATP: Contextually Adaptive Token Pruning for Efficient and Enhanced Multimodal In‑Context Learning}
\author{
    Yanshu Li\textsuperscript{\rm 1},
    Jianjiang Yang\textsuperscript{\rm 2},
    Zhennan Shen\textsuperscript{\rm 1},
    Ligong Han\textsuperscript{\rm 3},
    Haoyan Xu\textsuperscript{\rm 4},
    Ruixiang Tang\textsuperscript{\rm 5}\thanks{Corresponding Author.}
}
\affiliations{
    \textsuperscript{\rm 1}Brown University\\
    \textsuperscript{\rm 2}University of Bristol\\
    \textsuperscript{\rm 3}MIT-IBM Watson AI Lab\\
    \textsuperscript{\rm 4}University of Southern California\\
    \textsuperscript{\rm 5}Rutgers University\\
    
    yanshu\_li1@brown.edu, ruixiang.tang@rutgers.edu
    
}

\usepackage{bibentry}

\begin{document}

\maketitle

\begin{abstract}
Modern large vision-language models (LVLMs) convert each input image into a large set of tokens that far outnumber the text tokens. Although this improves visual perception, it also introduces severe image token redundancy. Because image tokens contain sparse information, many contribute little to reasoning but greatly increase inference cost. Recent image token pruning methods address this issue by identifying important tokens and removing the rest. These methods improve efficiency with only small performance drops. However, most of them focus on single-image tasks and overlook multimodal in-context learning (ICL), where redundancy is higher and efficiency is more important. Redundant tokens weaken the advantage of multimodal ICL for rapid domain adaptation and lead to unstable performance. When existing pruning methods are applied in this setting, they cause large accuracy drops, which exposes a clear gap and the need for new approaches. To address this, we propose Contextually Adaptive Token Pruning (CATP), a training-free pruning method designed for multimodal ICL. CATP uses two stages of progressive pruning that fully reflect the complex cross-modal interactions in the input sequence. After removing 77.8\% of the image tokens, CATP achieves an average performance gain of 0.6\% over the vanilla model on four LVLMs and eight benchmarks, clearly outperforming all baselines. At the same time, it improves efficiency by reducing inference latency by an average of 10.78\%. CATP strengthens the practical value of multimodal ICL and lays the foundation for future progress in interleaved image-text settings.
\end{abstract}

\section{Introduction}

\label{sec:intro}
\begin{figure}[ht] 
  \centering                 
  \includegraphics[width=\columnwidth]{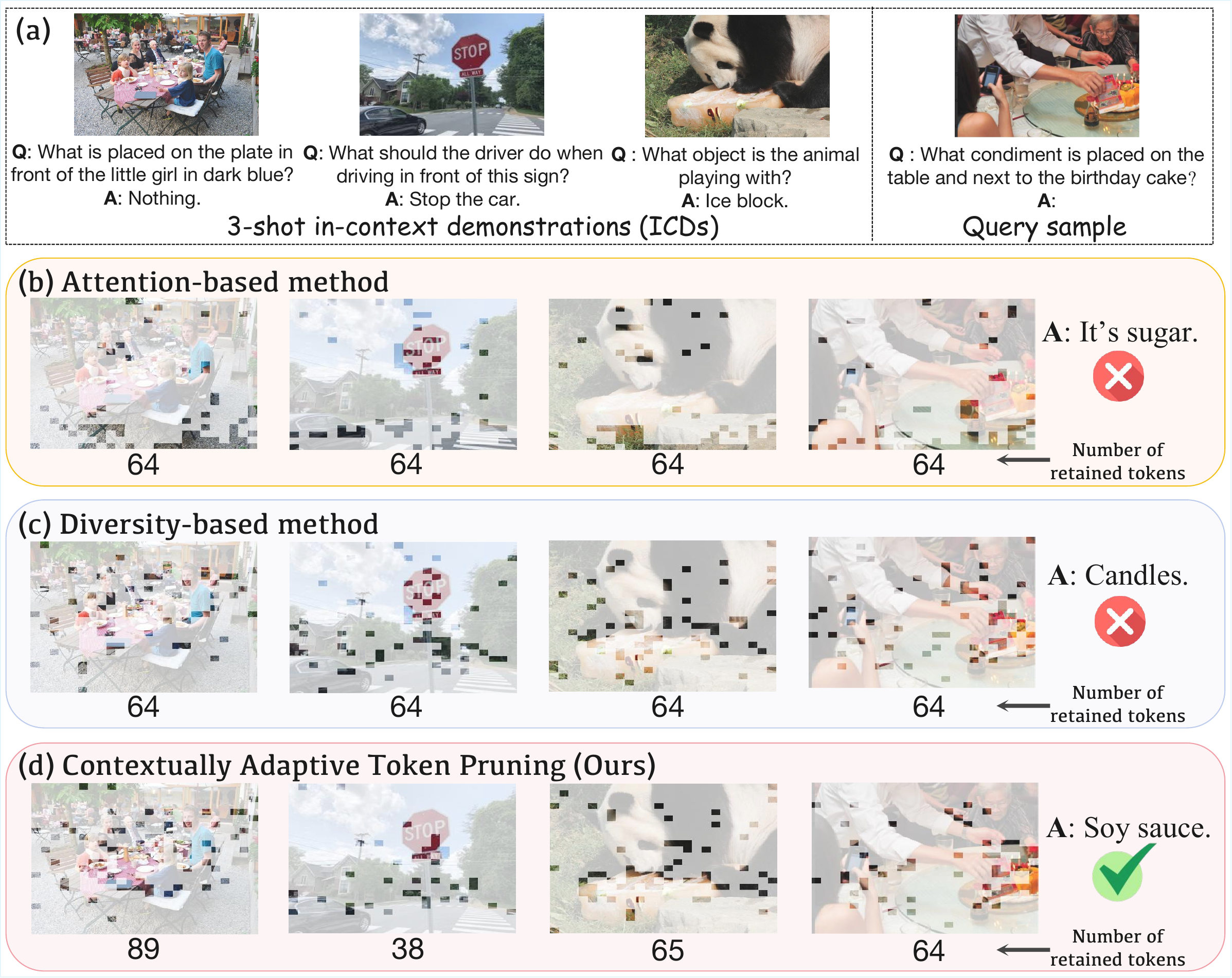}
  \caption{(a) Example of a 3-shot in-context sequence. (b-d) show the effects of three image token pruning methods on this sequence. Attention-based methods tend to keep tokens at the bottom of the image due to attention sinks in interleaved inputs. Diversity-based methods lack semantic guidance from the entire context. In contrast, our proposed method \textbf{CATP} accounts for the complex cross-modal interactions within the sequence, leading to superior results.}           
  \label{first}           
\end{figure}
Large vision-language models (LVLMs) have recently shown strong performance across a wide range of visual-language tasks \cite{vituning, intern1, qwen1}. To equip LVLMs with new domain capabilities while reducing the high cost of multimodal data preparation and training, multimodal in-context learning has become an important extension \cite{icl1,icl2}. It allows a model to adapt on the fly by adding a small set of exemplars, known as in-context demonstrations (ICDs), directly into the prompt. This paradigm provides a simple path toward task generalization without any parameter updates, making it increasingly attractive for real-world use \cite{icv2}.

Although multimodal ICL offers a user-friendly and rapid solution, its input characteristics often impede the desired effects. As shown in Figure \ref{first}(a), every ICD and the subsequent query sample include an image, so the image token redundancy that is already a bottleneck in single-image tasks becomes even more acute. For instance, LLaVA-Next \cite{LLaVAnext} converts each image into 576 tokens. On VizWiz \cite{VizWiz}, a 2-shot ICL run requires 3.2 times the computation of single-image inference and an impressive 14.3 times that of text-only inference. The resulting latency and memory footprint conflict with the efficiency users expect. Moreover, the interleaved format introduces additional issues. Multimodal ICL involves several kinds of cross-modal interactions, and many uninformative tokens can complicate this process \cite{pos, taco}. They prevent the model from focusing on the visual regions in each image that match the paired text. They also make it harder to locate the key information provided by the user based on the query sample \cite{vega,comm}.

To ease this serious redundancy, training‑free image token pruning has become a promising direction. Its core idea is to define an importance criterion that measures the contribution of each image token during inference \cite{pruning1,pruning2}. By retaining only the most informative tokens and removing or merging the rest, latency is effectively reduced with minimal accuracy loss. Existing pruning strategies fall mainly into two categories. (1) \textbf{Diversity‑based} methods operate after the vision encoder and projector but before the decoder, when image tokens have been projected into the decoder’s representation space yet have not interacted with texts \cite{div1,div2}. These methods measure feature similarity among image tokens to penalize redundancy and encourage diversity, as exemplified by DivPrune \cite{div}. (2) \textbf{Attention‑based} methods act inside the LLM decoder, where visual and textual embeddings are processed together \cite{att1,att2,att3}. They estimate the importance of each image token by the amount of attention it receives from other tokens, as exemplified by FastV \cite{fastv}.

\begin{figure}[t] 
  \centering                 
  \includegraphics[width=\columnwidth]{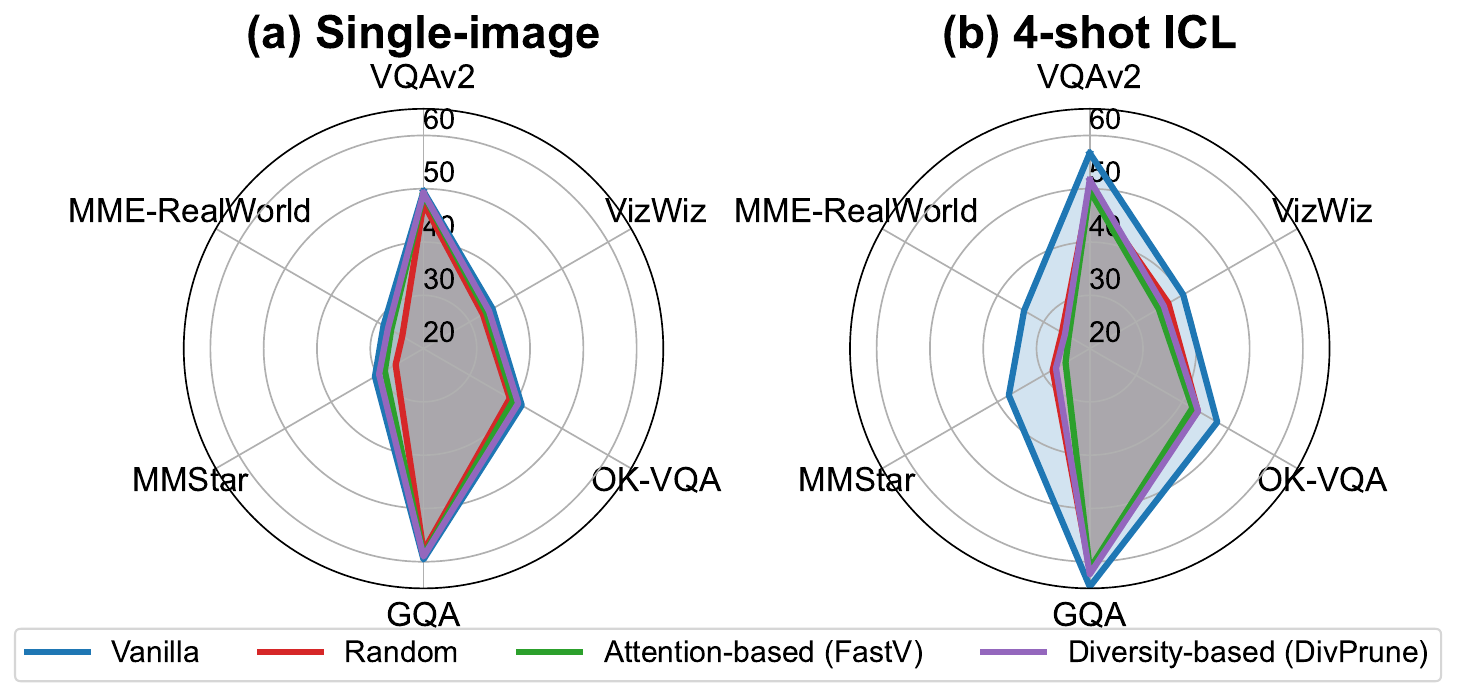}
  \caption{Performance comparison of three token pruning methods under a 77.8\% per‑image pruning ratio, shown for single‑image versus 4‑shot ICL across six benchmarks.} 
  \label{radar}           
\end{figure}

Despite being able to reduce up to 90\% of the image tokens with less than a 5\% drop in performance, the above two pruning methods are developed and evaluated only in single‑image tasks and thus ignore the much more demanding multi‑image scenarios, especially interleaved image-text ICL. When we adapt FastV and DivPrune to multimodal ICL, their shortcomings become clear, as shown in the visualization of Figure \ref{first} and the results on LLaVA‑Next‑7B in Figure \ref{radar}. Both methods incur greater performance degradation in 4‑shot ICL compared to the single-image setting, and in many cases even underperform random pruning. This highlights a non‑negligible gap between single‑image tasks and multimodal ICL. To investigate this gap, we conduct further ICL experiments with both pruning methods under varied setups and obtain two key findings. First, in multimodal ICL, diversity-based methods fail to perform fine-grained pruning as they lack information from other images and texts, and thus cannot capture diverse cross-modal interactions. Second, while layer choice matters for attention‑based pruning, attention shifts \cite{limi1} in every image bias all its variants toward text‑adjacent tokens, and the bias compounds to reduce accuracy. To this end, we question that

\textit{``Can we identify the image tokens that contribute the most to building a complete and effective in-context sequence amid complex cross‑modal interactions?''}

To approach this target, we propose Contextually Adaptive Token Pruning (CATP), a training-free image token pruning method designed for multimodal ICL. CATP removes a ratio $R$ of the total image tokens in an in-context sequence. Based on our analysis of existing methods, CATP is divided into two stages. Stage 1 is applied between the projector and the decoder, and is tailored to the characteristics of multimodal ICL. It evaluates the token importance in two dimensions. The first is semantic alignment with the paired text, which ensures that key visual regions are retained as a foundation for effective ICL. The second is feature diversity, which prevents the tokens that matter more to the entire sequence from being discarded. Stage 2 acts on the shallow decoder layers. It adopts a novel progressive adaptation strategy that first treats all ICD image tokens as an integral context. By combining semantic relevance with layer-wise attention differences, it precisely identifies the most important context tokens. In the following layer, it prunes the query image tokens based on their semantic relevance to this distilled context, ensuring effective ICL after pruning. Extensive experiments show that CATP is both effective and generalizable in multimodal ICL. It improves efficiency and performance at the same time and provides clear advantages over all existing baselines.

The contributions of this paper are summarized below:
\begin{itemize}
    \item  By testing two mainstream image token pruning methods in the multimodal ICL setting, we reveal that their effectiveness degrades due to the interleaved image-text nature of the input, which amplifies their limitations.
    \item We propose Contextually Adaptive Token Pruning (CATP), the first image token pruning method designed for multimodal ICL, filling the important gap. CATP is training-free and can capture multiple types of interaction to select the tokens most critical for overall ICL.
    \item Experiments on four LVLMs and eight benchmarks show that, while all other baselines lead to performance degradation, CATP not only improves the efficiency of multimodal ICL but also enhances its performance.
\end{itemize}

\section{Related Work}
\begin{figure*}[ht] 
  \centering                 
  \includegraphics[width=0.9\textwidth]{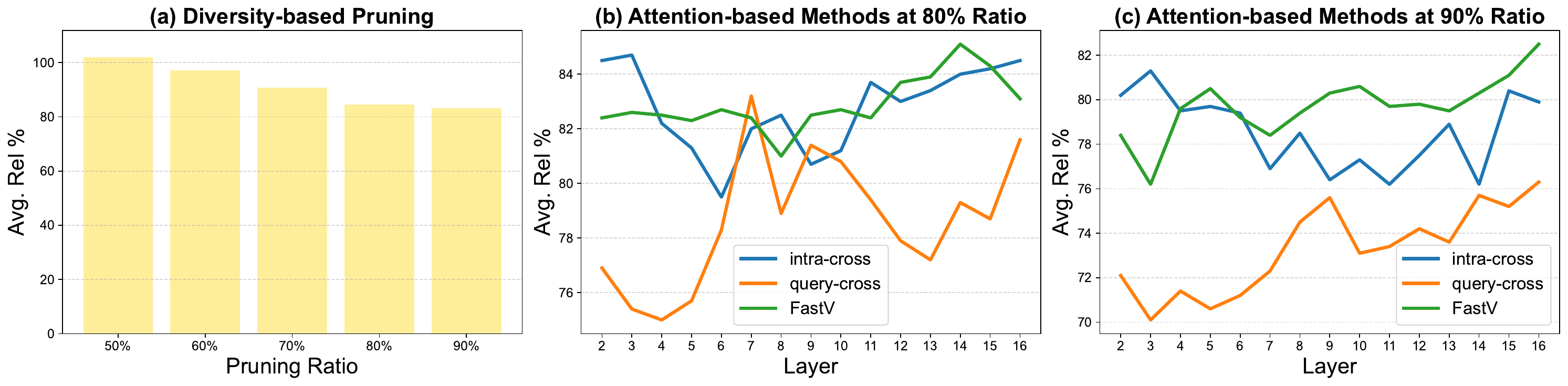}
  \caption{Average relative performance of pruning methods in 4-shot ICL: (a) showing diversity‑based pruning at 50\%-90\% ratios; (b-c) showing three different attention‑based pruning methods applied to various layers at 80\% and 90\% ratios.} 
  \label{analy}           
\end{figure*}

\textbf{Large Vision-language Models (LVLMs).} The rapid rise of large language models (LLMs) gives birth to LVLMs \cite{llava, onevision, intern1, qwen1}. A typical LVLM comprises \textbf{three} key modules: a vision encoder, a projector, and an LLM decoder. Although this architecture is generally effective, it has been found to be inefficient \cite{token1,token2}. Input images usually occupy a large number of tokens, and these tokens carry much sparser information than text tokens, creating severe image token redundancy \cite{redun1,redun2}. For example, LLaVA‑1.5 \cite{llava} converts each image into 576 tokens, while LLaVA‑Next \cite{LLaVAnext} and Qwen2.5-VL \cite{qwen} increase the count up to several thousand tokens per image to handle high‑resolution inputs.

\textbf{In-context Learning (ICL).} ICL refers to a model’s ability to solve new tasks by conditioning on a sequence of in-context demonstrations (ICDs) without updating any parameters \cite{iclr1, iclr2, iclr3}. Because of its convenience, ICL has been widely used in NLP and extended to the multimodal setting. Current-generation LVLMs, including the InternVL and QwenVL series, already treat ICL as a core capability, which highlights its strong practical value \cite{intern1,qwen1}. However, the ideal efficiency of ICL is often reduced by image token redundancy, and the complex interactions in multimodal ICL make this problem even harder to address.

\textbf{Image Token Pruning.} Researchers have begun to tackle token redundancy by image token pruning. FastV \cite{fastv} first proposes to prune image tokens based on the attention they obtain in an early decoder layer. Following FastV, more attention‑based methods have appeared \cite{att1,att2}. VTW \cite{vtw} proposes a token withdrawal strategy, removing all image tokens after a chosen layer. FitPrune \cite{fit} further leverages both cross‑attention and self‑attention of image tokens in every layer to measure their importance. Meanwhile, diversity-based methods argue that semantic diversity among image tokens is a better importance criterion than attention scores \cite{div1,div2}. DivPrune \cite{div} performs pruning before image tokens enter the decoder and frames it as a Max-Min Diversity Problem. Although all these methods are effective, they were designed for single-image tasks and overlook multimodal ICL. We find that their known limitations \cite{pruning2,limi1} become even more pronounced in this setting.

\section{Method}

\subsection{Preliminary and Motivation}
Consider an LVLM parameterized by $\theta$ and an $n$-shot in-context sequence $ICS$ that contains $n$ ICDs with both image and text, plus one query sample. We denote the sequence as:
\begin{equation}
ICS=\{(I_{1},T_{1}),\ldots,(I_{n},T_{n}),(I_{q},T_{q})\},
\end{equation}
where $I_{q}$ and $T_{q}$ are the image and text of the query sample. In the prefilling stage, each $i$-th image in $ICS$ is first passed through an image encoder $f_{\theta}^{v}$ and then a projector $g_{\theta}$, producing a set of image token embeddings:
\begin{equation}
    X^{I}_i=g_{\theta}(f_{\theta}^{v}(I_{i})) \in \mathbb{R}^{S^{I}_{i} \times D},
\end{equation}
where $D$ denotes the dimension of the LLM decoder $\phi^{\theta}$ and $S^{I}_{i}$ is the number of tokens produced from each image. This number varies across LVLMs and is usually user-configurable. \textbf{Diversity-based} pruning methods are typically applied at this stage to reduce $S^{I}_{i}$. Meanwhile, each $i$-th text segment is converted into a set of text token embeddings $X^{T}_{i}\in\mathbb{R}^{S^{T}_{i} \times D}$ by $\phi^{\theta}$. After inserting the image tokens at their corresponding positions, we obtain the input token sequence:
\begin{equation}
    X=(X_{sys},X^{I}_1,\,X^{T}_1,\,\dots,\,X^{I}_n,\,X^{T}_n,\,X^{I}_{n+1},\,X^{T}_{n+1}),
\end{equation}
where $X_{sys}$ denotes the system prompt and the query sample is indexed as $n$+$1$ for unified notation. The sequence $X$ is then fed into $\phi^{\theta}$ and processed by an $N$-layer decoder to produce the final answer. Each layer contains a multi-head attention module (MHA) and a feedforward MLP. The $h$-th head in the $l$-th layer maps the hidden states of $X_{ICS}$ to queries $Q_{l,h}$, keys $K_{l,h}$, and values $V_{l,h}$ by linear transformations, and the head's attention weight matrix can be represented as:
\begin{equation}
\label{attneq}
\mathbf{A}_{l,h}
  =
      {softmax}(\frac{Q_{l,h}\,K_{l,h}^{\top}+\mathbf{M}}
           {\sqrt{D_k}})\in \mathbb{R}^{S \times S}
\end{equation}
where $S$ is the length of $X$ and $\mathbf{M}$ is the causal mask. $D_k$ is the head dimension of $\phi^{\theta}$. $K_{l,h}$ and $V_{l,h}$ are stored in the KV cache during the prefilling stage to prepare for subsequent decoding. \textbf{Attention-based} pruning methods are applied here to leverage $\mathbf{A}_{l,h}$ to discard less informative image tokens, preventing them from entering later layers. Both methods aim to identify the tokens whose removal minimizes the adverse effects on $\theta$'s decoding, given a specific pruning ratio.

To further examine the gap revealed in Figure \ref{radar}, we conduct additional 4-shot ICL evaluations on the six benchmarks using LLaVA-Next-7B and report the average relative performance (Avg. Rel) compared with the vanilla model. We first adapt DivPrune to ICL with pruning ratios ranging from 50\% to 90\%. As Figure \ref{analy}(a) shows, it performs well at relatively low ratios and at 50\% it even outperforms the vanilla model. When the ratio increases from 60\% to 70\%, its performance decreases sharply and continues to decline. These results indicate that diversity-based methods \textbf{fail to deliver fine-grained pruning in multimodal ICL}. In these methods, each image token interacts only with tokens from the same image, so no information from other images or texts is considered. Meanwhile, multimodal ICL involves much richer cross‑modal interactions than single-image tasks, making this limitation more pronounced. Nonetheless, they remain useful as an \textbf{initial filter} before the decoder.

Next, we evaluate FastV by pruning 80\% and 90\% of image tokens at different decoder layers ranging from 2 to 16. FastV ranks tokens by the total attention they receive from all other tokens in a chosen layer. Considering the property of multimodal ICL, we introduce two alternatives: \textit{intra‑cross} measures the attention that each image token receives from all text tokens within its own image-text pair, and \textit{query‑cross} measures the attention that each image token receives from all tokens in the query sample. Intra‑cross performs best in very shallow layers, such as layer 2, where the model is focused on feature perception and alignment. As the layer index increases, the effectiveness of intra‑cross declines while FastV becomes superior. Query‑cross shows a sharp rise in the shallow layers, roughly layers 7 to 10, indicating that after perception, the LVLM shifts to query‑guided reasoning.

However, none of the three methods reaches the desired performance in ICL. We attribute this to \textbf{attention shifts} that cause the model to focus excessively on image tokens positioned closer to the text segments \cite{limi1}. In interleaved settings, these shifts accumulate and make the attention signal even less indicative of true importance, as illustrated in Figure \ref{first}(b) and Appendix 2. Hence, \textbf{statically relying on the attention assigned to image tokens is suboptimal} for pruning in multimodal ICL. 
\begin{figure}[t] 
  \centering                 
  \includegraphics[width=0.95\columnwidth]{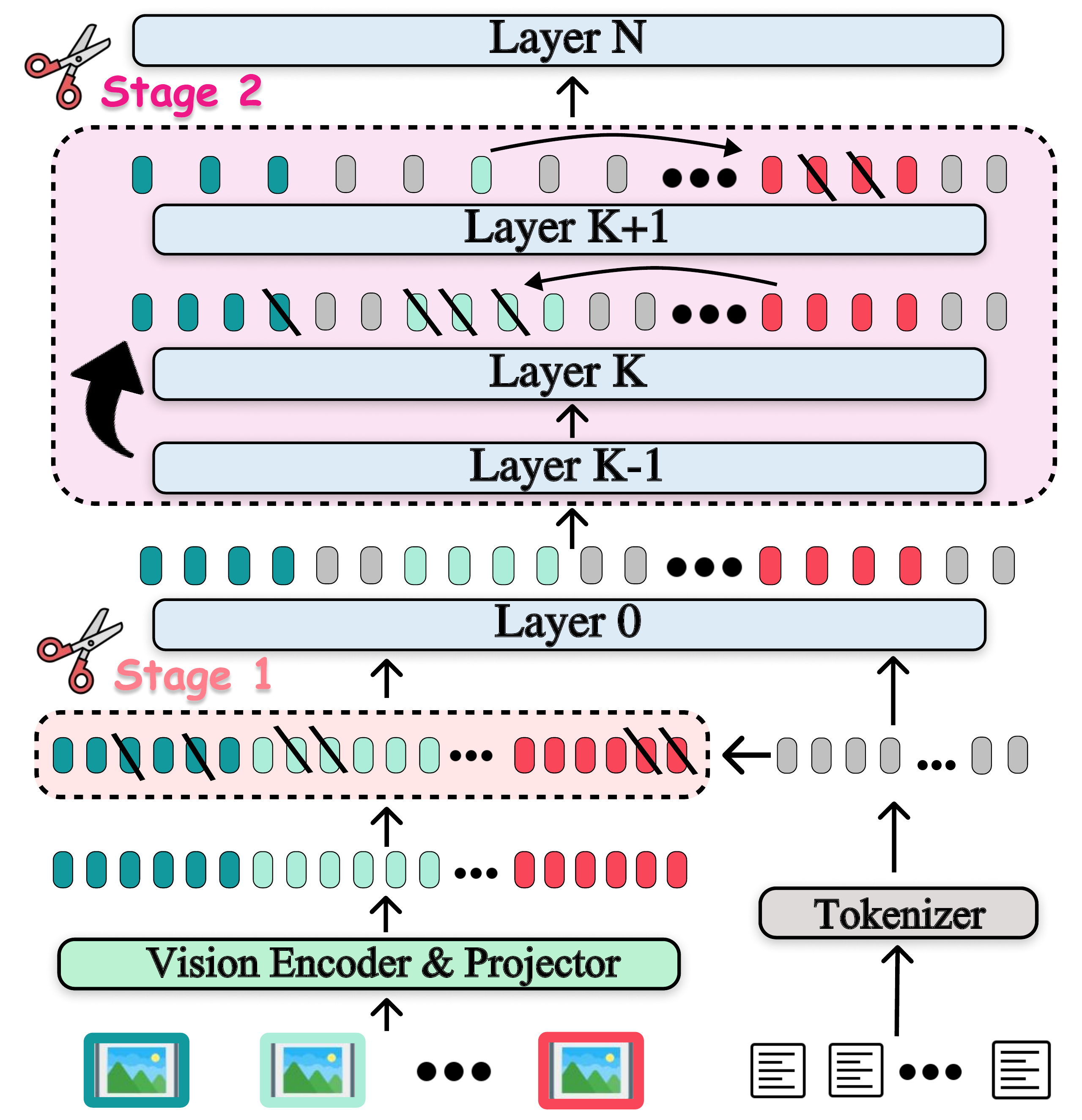}
  \caption{An overview pipeline of CATP. The version with more details is provided in Appendix 1.}             
  \label{main}           
\end{figure}

\subsection{Contextually Adaptive Token Pruning}
\subsubsection{Overview}
To bridge the gap left by existing pruning methods in multimodal ICL, we propose Contextually Adaptive Token Pruning (CATP). Figure \ref{main} outlines its pipeline. Given a target pruning ratio $R$, CATP retains only $(1-R)$ of the total image tokens for decoding, which effectively reduces inference latency while keeping ICL performance nearly unaffected. Guided by the preceding analysis, CATP proceeds in two stages to deliver progressive and thorough pruning. Each stage removes $R/2$ of the total image tokens in $X$. Stage 1 is applied between the projector and the decoder. It measures the importance of image tokens using feature distribution diversity and alignment with their paired text. Stage 2 operates in two shallow decoder layers. It first prunes the ICD image tokens by combining inter-layer attention differences with semantic relevance, and then prunes the query sample’s image tokens in the following layer.

\subsubsection{Stage 1}
Due to the multiple images in the input $ICS$, the converted token sequence $X$ is lengthy, causing the decoder to be hampered during prefilling and disrupting the intended attention flow. Thus, Stage 1 of CATP is performed directly after the projector to prune the least informative image tokens before they reach the decoder, maximizing overall precision and efficiency. We design a targeted strategy to keep the image tokens entering the decoder truly informative. 

In multimodal ICL, each image is paired with text, so semantically aligned tokens are crucial for reasoning. However, the whole ICL process can depend more on features that interact with other ICDs or the query sample, such as complementary cross-sample cues. Thus, we measure the importance of this stage in two dimensions. The first dimension assesses the \textbf{alignment} between each image token and its paired text. The advantage of considering alignment at this stage is two-fold. After reducing the number of tokens, it promotes correct attention in the decoder by easing attention shifts \cite{cama}. It also captures a key interaction in multimodal ICL that occurs within each image-text pair. This interaction is hard to isolate in the decoder as the forward pass merges semantics. The second assesses the \textbf{diversity} of the information carried by the image tokens. This keeps the retained tokens close to the feature distribution of the complete set and prevents tokens that contribute to the entire sequence from being removed. This matches the diversity-based methods, which are found to be suitable for initial filtering. To unify the two dimensions, we frame the task as a \textbf{weighted submodular maximization} problem. 

Formally, for each $i$-th visual embedding set $X^{I}_{i}$, our goal is to select an optimal subset $Y^{*}_{i}\subseteq X_{i}^{I}$ that maximizes a composite objective function $\mathcal{F}(Y_{i})$. This function is a weighted sum of the alignment and diversity scores. The optimization problem is defined as:
\begin{equation}
    Y_i^* = \arg\max_{\substack{Y_i \subseteq X_i^I,\\ |Y_i| = \left \lfloor S_i^I \cdot (1 - (R/2))\right \rfloor}} 
\big( \mathcal{F}_{\text{div}}(Y_i) + \lambda_1 \cdot \mathcal{F}_{\text{align}}(Y_i) \big),
\label{stage1}
\end{equation}
where $\lambda_1$ is a hyperparameter that balances the two objectives. The alignment function $\mathcal{F}_{\text{align}}(Y_i)$ quantifies how well the selected image tokens align with the semantics of their paired text segment $X_{i}^{T}$. We first compute a single vector representing the $i$-th text segment's semantics, $\bar{\mathbf{v}}_{i}$, by averaging all embeddings in $X_{i}^{T}$. Then, we obtain $\mathcal{F}_{\text{align}}(Y_i)$ using the similarity score between each image token $x_{i,j}\in X^{I}_{i}$ and $\bar{\mathbf{v}}_{i}$:

\begin{equation}
\mathcal{F}_{\text{align}}(Y_i) = \sum_{x_{i,j} \in Y_i}\hat{\text{sim}}(x_{i,j},\bar{\mathbf{v}}_{i}),
\end{equation}
where $\hat{\text{sim}}(\cdot)$ denotes the cosine similarity rescaled into the range $[0,1]$. The diversity function $\mathcal{F}_{\text{div}}(Y_i)$ is modeled with a facility location objective that drives the tokens in $Y_i$ to represent as diverse a set of features as possible. This function is a monotone submodular set function, measuring the total similarity between all tokens in the ground set $X^I_i$ and their closest token in the chosen subset $Y_i$:

\begin{equation}
\mathcal{F}_{\text{div}}(Y_i) = \sum_{x \in X^I_i} \max_{y \in Y_i} \hat{\text{sim}}(y, x).
\end{equation}

Since $\mathcal{F}_{\text{div}}(Y_i)$ is submodular and $\mathcal{F}_{\text{align}}(Y_i)$ is a modular function, their linear combination in Eq. \ref{stage1} remains submodular. This property allows us to efficiently find a near-optimal pruning strategy using a standard greedy algorithm, making stage 1 a lightweight and effective pre-decoder filter.

\subsubsection{Stage 2}
After Stage 1, the decoder receives a pruned token sequence whose image tokens equal $(1-R/2)$ of those in $X$. Only inside the decoder can we capture the complex interactions in multimodal ICL that go beyond image-text alignment via hidden states and attention patterns. Figure \ref{analy}(b-c) shows the sharp rises of query-cross in the shallow layers, indicating that LVLM begins to rely more on the query sample to locate key context features at this depth, making query-context interactions more salient. Thus, we apply Stage 2 here. As static single-layer attention fails to measure importance, we propose a novel strategy, \textbf{progressive adaptation}.

We set the starting layer index of progressive adaptation to $K$. At this layer, we treat all ICDs as an integral context and prune their image tokens altogether until the overall ratio $R$ is reached. Thus, unlike existing methods, the pruning ratio inside each ICD is not fixed. It adapts to the importance of each ICD to the context as a whole. Because not every ICD contributes equally to ICL and some are even detrimental, this strategy preserves post-pruning performance. We use the query sample as a lens to prune the context. First, we compute a focused query vector $\bar{\mathbf{v}}_{\text{q}}^K$ by applying average pooling to the hidden states of all tokens in the query sample at layer $K$. We then score each context image token $c$ by comprehensively considering its relevance to the query and its dynamic growth in importance This growth is measured by the change in attention it receives from the query sample's last token $t_{ql}$ between layers:
\begin{equation}
    \small
    \Delta \mathcal{A}(c) = \max\left(0, \mathbf{A}_K[idx_{ql},idx_c] - \mathbf{A}_{K-1}[idx_{ql},idx_c]\right),
\end{equation}
where $\mathbf{A}_l$ is the $l$-th layer's attention weight matrix averaged over all heads. $idx_{ql}$ and $idx_c$ denote the token
index of $t_{ql}$ and $c$, respectively. The full importance score is defined as: 
\begin{equation}
    \mathcal{S}_{\text{context}}(c) = \Delta \mathcal{A}(c)+ \lambda_2 \cdot \mathrm{sim}(\mathbf{h}_c^K, \bar{\mathbf{v}}_{\text{q}}^K),
\end{equation}
where $\mathbf{h}_c^{K}$ denotes the hidden states of $c$ at layer $K$ and $\lambda_2$ is a hyperparameter. We globally rank all context image tokens by this score and prune them in ascending order. This step simultaneously distills the large context, allowing the LVLM to conduct more targeted and in-depth reasoning. Next, at layer $K+1$ we perform context-guided pruning on the query sample’s image tokens. We apply average pooling to the hidden states of the pruned context at this layer to obtain a distilled context vector $\bar{\mathbf{v}}_c^{K+1}$. To maintain a focused mechanism, the importance of each query image token $q$ is only determined by its relevance to the distilled context:
\begin{equation}
     \mathcal{S}_{\text{query}}(q) = \mathrm{sim}(\mathbf{h}_q^{K+1}, \bar{\mathbf{v}}_{\text{c}}^{K+1}),
\end{equation}
where $\mathbf{h}_q^{K+1}$ is the hidden states of $q$ at layer $K+1$. After pruning the query sample’s image tokens with this score, CATP is complete. Only $1-R$ of the total image tokens in $X$ proceeds to layer $K+2$ and all subsequent inference.
\section{Experiments}

\begin{table*}[!ht]
    \centering
    \setlength{\tabcolsep}{3.5pt}
    \footnotesize
    \centering
    \scalebox{0.9}{
    \begin{tabular}{c | c c c c c c c c | c}
        \toprule[1.5pt]
        \textbf{Method} & \textbf{VQAv2} & \textbf{VizWiz} & \textbf{OK-VQA} & \textbf{GQA} & \textbf{Hatefulmemes} & \textbf{MMStar} & \textbf{MME-RealWorld} & \textbf{VL-ICL} & {\textbf{Avg. Rel}} \\
        \hline
        \rowcolor{lightgray}
         LLaVA-Next-7B& \multicolumn{9}{c}{\textit{Upper Bound 100\%}}\\
        Vanilla & 56.7 & 40.2 & 47.6 & 65.7 & 66.8 & 37.5 & 34.2 & 26.8 & 100.0\% \\
        \hline
        \rowcolor{lightgray}
         LLaVA-Next-7B& \multicolumn{9}{c}{\textit{Pruning Ratio 66.7\%}}\\
        Random & 52.3 & 37.4 & 44.9 & 63.4 & 64.5 & 30.9 & 28.8 & 22.7 & 90.5\% \\
        FastV \texttt{\scriptsize{(ECCV24)}} & 52.1 & 35.8 & 43.7 & 63.0 & 64.1 & 27.9 & 26.9 & 19.8 & 86.4\%\\
        HiRED \texttt{\scriptsize{(AAAI25)}} & 53.8 & 37.6 & 45.8 & 63.9 & 64.7 & 30.8 & 31.0 & 23.2 & 92.3\% \\
        FitPrune \texttt{\scriptsize{(AAAI25)}} & 51.0 & 36.7 & 44.5 & 62.3 & 63.6 & 28.4 & 28.1 & 21.7 & 88.0\% \\
        VTW \texttt{\scriptsize{(AAAI25)}} & 52.0 & 36.1 & 44.8 & 61.3 & 64.7 & 32.6 & 31.5 & 22.3 & 91.0\% \\
        DivPrune \texttt{\scriptsize{(CVPR25)}} & 53.5 & 37.0 & 43.7 & 63.5 & 64.3 & 29.2 & 30.8 & 23.4 & 90.8\% \\
        SparseVLM \texttt{\scriptsize{(ICML25)}} & 52.5 & 37.2 & 44.8 & 63.4 & 64.5 & 31.0 & 28.3 & 20.5 & 89.3\% \\
        PLPHP \texttt{\scriptsize{(2025.2)}} & 54.7 & 39.5 & 46.9 & 65.5 & 66.3 & 36.2 & 33.4 & 25.0 & 97.5\% \\
        CATP (Ours) & \textbf{57.1} & \textbf{40.6} & \textbf{47.9} & \textbf{66.4} & \textbf{67.0} & \textbf{37.6} & \textbf{34.0} & \textbf{27.3} & \textbf{100.7\%} \\
        \hline
        \rowcolor{lightgray}
         LLaVA-Next-7B& \multicolumn{9}{c}{\textit{Pruning Ratio 77.8\%}}\\
         Random & 49.8 & 37.0 & 43.4 & 63.2 & 63.9 & 28.0 & 26.3 & 18.4 & 85.4\% \\
        FastV \texttt{\scriptsize{(ECCV24)}} & 49.7 & 34.9 & 42.3 & 62.4 & 63.5 & 25.3 & 24.6 & 17.7 & 82.4\% \\
        HiRED \texttt{\scriptsize{(AAAI25)}} & 52.4 & 36.9 & 44.5 & 62.7 & 64.0 & 28.5 & 28.9 & 21.6 & 88.8\% \\
        FitPrune \texttt{\scriptsize{(AAAI25)}} & 50.1 & 35.6 & 43.7 & 62.0 & 63.2 & 27.3 & 27.0 & 18.2 & 84.7\% \\
        VTW \texttt{\scriptsize{(AAAI25)}} & 50.9 & 33.6 & 41.7 & 61.5 & 62.7 & 30.2 & 28.9 & 21.8 & 86.9\% \\
        DivPrune \texttt{\scriptsize{(CVPR25)}} & 51.7 & 36.0 & 43.5 & 63.3 & 63.4 & 27.3 & 26.0 & 20.9 & 86.3\% \\
        SparseVLM \texttt{\scriptsize{(ICML25)}} & 49.5 & 35.8 & 42.7 & 62.6 & 64.0 & 26.8 & 25.4 & 18.3 & 83.9\% \\
        PLPHP \texttt{\scriptsize{(2025.2)}} & 54.0 & 39.1 & 46.0 & 64.8 & 65.5 & 35.9 & 32.7 & 24.1& 95.9\% \\
        CATP (Ours) & \textbf{56.8} & \textbf{40.0} & \textbf{47.8} & \textbf{65.9} & \textbf{66.7} & \textbf{37.2} & \textbf{34.4} & \textbf{27.0} & \textbf{100.1\%} \\
        \hline
        \rowcolor{lightgray}
         LLaVA-Next-7B& \multicolumn{9}{c}{\textit{Pruning Ratio 89.9\%} }\\
         Random & 46.3 & 34.6 & 42.3 & 60.1 & 61.5 & 25.8 & 23.9 & 16.2 & 79.9\% \\
        FastV \texttt{\scriptsize{(ECCV24)}} & 46.0 & 33.5 & 41.8 & 59.3 & 61.7 & 23.5 & 23.2 & 16.5 & 78.4\% \\
        HiRED \texttt{\scriptsize{(AAAI25)}} & 48.6 & 35.0 & 42.9 & 61.9 & 62.7 & 26.4 & 26.0 & 19.2 & 83.6\% \\
        FitPrune \texttt{\scriptsize{(AAAI25)}} & 46.1 & 34.6 & 42.5 & 61.1 & 62.0 & 25.3 & 22.7 & 16.4 & 79.7\% \\
        VTW \texttt{\scriptsize{(AAAI25)}} & 47.6 & 32.8 & 37.4 & 60.1 & 61.5 & 26.3 & 25.9 & 18.7 & 80.4\% \\
        DivPrune \texttt{\scriptsize{(CVPR25)}} & 48.2 & 34.8 & 43.0 & 62.3 & 62.5 & 25.7 & 25.0 & 20.1 & 83.4\% \\
        SparseVLM \texttt{\scriptsize{(ICML25)}} & 46.2 & 31.3 & 41.6 & 61.3 & 61.9 & 25.4 & 24.3 & 17.0 & 79.4\% \\
        PLPHP \texttt{\scriptsize{(2025.2)}} & 53.4 & 39.0 & 45.6 & 65.0 & 65.2 & 35.8 & 32.1 & 23.3 & 95.0\% \\
        CATP (Ours) & \textbf{56.3} & \textbf{39.7} & \textbf{47.2} & \textbf{65.4} & \textbf{66.3} & \textbf{36.7} & \textbf{34.0} & \textbf{25.5} & \textbf{98.6\%} \\
        \bottomrule[1.5pt]
    \end{tabular}}

	\caption{4-shot performance comparison of different pruning methods on LLaVA-Next-7B under three pruning ratios. \textbf{Avg. Rel} denotes the average percentage of performance relative to the vanilla model.}
    \label{tab:main1}
\end{table*}

\begin{table*}[ht]
    \centering
    \setlength{\tabcolsep}{3.5pt}
    \footnotesize
    \centering
    \scalebox{0.9}{
    \begin{tabular}{c | c c c c c c c c | c}
        \toprule[1.5pt]
        \textbf{Method} & \textbf{VQAv2} & \textbf{VizWiz} & \textbf{OK-VQA} & \textbf{GQA} & \textbf{HatefulMemes} & \textbf{MMStar} & \textbf{MME-RealWorld} & \textbf{VL-ICL} & {\textbf{Avg. Rel}} \\
        \hline
        \rowcolor{lightgray}
        \multicolumn{10}{c}{LLaVA-Next-13B }\\
         Vanilla & 60.3 & 42.7 & 49.9 & 70.1 & 69.8 & 40.2 & 39.4 & 28.5 & 100\%\\
         Random & 53.2 & 39.4 & 44.0 & 65.3 & 65.7 & 33.4 & 34.1 & 23.2 & 88.4\% \\
        FastV \texttt{\scriptsize{(ECCV24)}} & 51.9 & 38.5 & 44.2 & 65.1 & 66.0 & 32.3 & 33.3 & 21.9 & 86.7\% \\
        DivPrune \texttt{\scriptsize{(CVPR25)}} & 54.5 & 40.1 & 46.3 & 65.4 & 66.8 & 34.5 & 34.9 & 24.8 & 90.9\% \\
        PLPHP \texttt{\scriptsize{(2025.2)}} & 58.7 & 42.3 & 49.3 & 69.8 & 69.8 & 39.3 & 38.5 & 23.0 & 96.4\% \\
        CATP (Ours) & \textbf{60.5} & \textbf{43.2} & \textbf{50.1} & \textbf{70.5} & \textbf{70.1} & \textbf{39.7} & \textbf{39.2} & \textbf{28.6} & \textbf{100.2\%} \\
        \hline
        \rowcolor{lightgray}
         \multicolumn{10}{c}{InternVL2.5-8B }\\
         Vanilla & 66.3 & 55.7 & 60.2 & 70.9 & 73.2 & 57.4 & 64.9 & 41.5 & 100\%\\
         Random & 59.4 & 49.3 & 55.8 & 68.5 & 71.5 & 49.1 & 57.8 & 35.9 & 90.8\% \\
        FastV \texttt{\scriptsize{(ECCV24)}} & 59.6 & 48.8 & 55.0 & 67.7 & 72.0 & 47.5 & 58.2 & 34.3 & 89.7\% \\
        DivPrune \texttt{\scriptsize{(CVPR25)}} & 60.8 & 50.5 & 54.7 & 67.3 & 70.2 & 50.9 & 60.3 & 35.7 & 91.5\% \\
        PLPHP \texttt{\scriptsize{(2025.2)}} & 65.8 & 50.0 & 56.5 & 70.3 & 73.4 & 56.5 & 64.3 & 40.6 & 97.2\% \\
        CATP (Ours) & \textbf{66.0} & \textbf{56.0} & \textbf{60.4} & \textbf{80.3} & \textbf{73.7} & \textbf{57.9} & \textbf{64.5} & \textbf{41.7} & \textbf{101.9\%} \\
        \hline
        \rowcolor{lightgray}
         \multicolumn{10}{c}{Qwen2.5VL-7B}\\
         Vanilla & 70.3 & 54.9 & 63.7 & 73.0 & 74.9 & 61.8 & 63.0 & 45.3 & 100\%\\
         Random & 60.3 & 47.2 & 60.8 & 70.0 & 69.7 & 55.8 & 56.6 & 37.4 & 89.9\% \\
        FastV \texttt{\scriptsize{(ECCV24)}} & 58.6 & 47.3 & 59.4 & 69.7 & 69.2 & 54.9 & 53.1 & 35.4 & 87.7\% \\
        DivPrune \texttt{\scriptsize{(CVPR25)}} & 61.2 & 47.0 & 62.6 & 72.3 & 73.5 & 56.4 & 56.3 & 40.1 & 92.2\% \\
        PLPHP \texttt{\scriptsize{(2025.2)}} & 67.3 & 53.6 & 63.5& 72.6 & 74.0 & 59.7 & 61.4 & 44.6& 98.0\% \\
        CATP (Ours) & \textbf{70.7} & \textbf{55.3} & \textbf{63.9} & \textbf{73.6} & \textbf{75.2} & \textbf{61.2} & \textbf{62.5} & \textbf{45.0} & \textbf{100.1\%} \\
        \bottomrule[1.5pt]
    \end{tabular}}

	\caption{4-shot performance comparison of different pruning methods on LLaVA-Next-13B, InternVL2.5-8B, and Qwen2.5VL-7B at a fixed pruning ratio of 77.8\% . \textbf{Avg. Rel} denotes the average percentage of performance relative to the vanilla model.}
    \label{tab:main2}
\end{table*}
\subsection{Setup}

\textbf{Benchmarks.} To validate CATP, we conduct extensive experiments on eight vision-language benchmarks. They include standard multimodal ICL evaluations \cite{icl3}: VQAv2 \cite{vqav2}, VizWiz \cite{VizWiz}, OK-VQA \cite{OK-VQA}, GQA \cite{GQA}, and HatefulMemes \cite{hateful}. Three latest and challenging multimodal benchmarks are also used: MMStar \cite{mmstar}, MME‑Realworld \cite{mme}, and VL‑ICL \cite{vlicl}.

\textbf{Baselines and models.} We compare CATP with the vanilla model, random pruning, and seven strong baselines. In addition to the previously introduced FastV, FitPrune, VTW, and DivPrune, we also evaluate HiRED \cite{att2}, SparseVLM \cite{att1}, and PLPHP \cite{plphp}. For methods relying on user‑specified hyperparameters, we test four configurations and report the \textbf{best} performance. We first evaluate each pruning method on LLaVA‑Next‑7B under a range of pruning ratios. We then report additional results at a fixed ratio on LLaVA‑Next‑13B, InternVL2.5‑8B, and Qwen2.5VL‑7B. In LLaVA-Next every image in the input sequence is uniformly converted into 576 tokens, whereas in InternVL2.5-8B and Qwen2.5VL-7B, the number of tokens per image depends on its original resolution and ranges from 576 to 1280. Appendix 3.1 provides further introductions to the benchmarks, baselines, models, and associated processing details.

\textbf{Implementation details.} All experiments are conducted on 4 NVIDIA H200 GPUs. For every benchmark the validation samples serve as query samples. Each query is paired with \textbf{four} ICDs randomly retrieved from the training split, producing a 4-shot sequence. For LLaVA-Next-7B, InternVL2.5-8B, and Qwen2.5VL-7B, we set $K=6$. For LLaVA-Next-13B, we set $K=10$. Across all LVLMs, we set $\lambda_1=0.7$ and $\lambda_2=0.6$. As CATP operates without the full attention matrices, it is compatible with FlashAttention \cite{flash}, which further boosts its efficiency.

\subsection{Main Results}
Table \ref{tab:main1} reports the performance of CATP under three commonly used pruning ratios. These are the levels at which pruning methods typically bring real efficiency gains. CATP achieves the best performance under all three settings. At both 66.7\% and 77.8\%, it enhances performance over the vanilla model. When the ratio reaches 89.9\%, CATP incurs only a modest 1.4\% drop. Meanwhile, CATP is effective across diverse LVLMs and benchmarks, as shown in Tables \ref{tab:main1} and \ref{tab:main2}. On four LVLMs and eight benchmarks, CATP consistently achieves the best performance. Notably, none of the seven SOTA baselines improve multimodal ICL, with some even performing worse than random pruning. In contrast, CATP achieves performance enhancement by capturing complex interactions and identifying the most important image tokens for the overall multimodal ICL process.

\begin{table}[ht]
    \centering
    \small
    \setlength{\tabcolsep}{1.3pt}
    \scalebox{0.9}{
    \begin{tabular}{c | c c c c}
        \toprule[1.5pt]
        \textbf{Method} & \textbf{FLOPs} & \textbf{Latency} & \textbf{KV Cache} & \textbf{Avg.~Rel} \\
        \hline
        \rowcolor{lightgray}
\multicolumn{5}{c}{LLaVA-Next-7B} \\
Vanilla & 20.84 & 3.82s & 1.12GB & 100\% \\
FastV        &  6.37 (69.4\%↓) & 3.25s (14.9\%↓) & 0.38GB (66.1\%↓) & 82.4\% \\
DivPrune     &  4.19 (79.9\%↓) & 4.03s (5.5\%↑)  & 0.30GB (73.2\%↓) & 86.3\% \\
PLPHP        &  5.33 (74.4\%↓) & 3.49s (8.6\%↓)  & 0.37GB (67.0\%↓) & 95.9\% \\
CATP  &  4.38 (79.0\%↓) & 3.19s (16.5\%↓) & 0.35GB (68.8\%↓) & \textbf{100.1\%} \\
\hline
\rowcolor{lightgray}
\multicolumn{5}{c}{LLaVA-Next-13B} \\
Vanilla & 39.94 & 5.24s & 2.03GB & 100\% \\
FastV        & 14.00 (64.9\%↓) & 4.83s (7.8\%↓)  & 0.51GB (74.9\%↓) & 89.7\% \\
DivPrune     &  9.38 (76.5\%↓) & 5.72s (9.2\%↑)  & 0.42GB (79.3\%↓) & 91.5\% \\
PLPHP        & 13.62 (65.9\%↓) & 4.95s (5.5\%↓)  & 0.55GB (72.9\%↓) & 97.2\% \\
CATP  & 10.15 (74.6\%↓) & 4.79s (8.6\%↓)  & 0.49GB (75.9\%↓) & \textbf{100.2\%} \\
\hline
\rowcolor{lightgray}
\multicolumn{5}{c}{InternVL2.5-8B} \\
Vanilla & 28.69 & 2.75s & 460.8MB & 100\% \\
FastV        &  9.63 (66.4\%↓) & 2.57s (6.5\%↓)  & 103.5MB (77.5\%↓) & 89.7\% \\
DivPrune     &  6.75 (76.5\%↓) & 2.90s (5.5\%↑)  &  82.3MB (82.1\%↓) & 91.5\% \\
PLPHP        &  8.10 (71.8\%↓) & 2.55s (7.3\%↓)  & 156.2MB (66.1\%↓) & 97.2\% \\
CATP  &  7.03 (75.5\%↓) & 2.53s (8.0\%↓)  &  97.6MB (78.8\%↓) & \textbf{101.9\%} \\
\hline
\rowcolor{lightgray}
\multicolumn{5}{c}{Qwen2.5VL-7B} \\
Vanilla & 27.43 & 2.31s & 215.0MB & 100\% \\
FastV        &  9.26 (66.2\%↓) & 2.11s (8.7\%↓)  &  47.5MB (77.9\%↓) & 87.7\% \\
DivPrune     &  6.18 (77.5\%↓) & 2.40s (3.9\%↑)  &  37.6MB (82.5\%↓) & 92.2\% \\
PLPHP        &  7.93 (71.1\%↓) & 2.19s (5.2\%↓)  &  75.2MB (65.0\%↓) & 98.0\% \\
CATP  &  6.48 (76.4\%↓) & 2.08s (10.0\%↓) &  41.7MB (80.6\%↓) & \textbf{100.1\%} \\
        \bottomrule[1.5pt]
    \end{tabular}}
    \caption{4-shot efficiency analysis of different pruning methods on four LVLMs. The results are averaged across 8 benchmarks at a ratio of 77.8\%. \textbf{Avg. Rel} denotes the average percentage of performance relative to the vanilla model.}
    \label{tab:eff}
\end{table}

\subsection{Efficiency Analysis} To demonstrate the efficiency of CATP, we conduct extensive comparative experiments on four LVLMs, reporting TFLOPs, inference latency, and KV Cache size. Following FastV, we compute the FLOPs of the multi-head attention and feedforward network modules as $4nd^2+2n^2d+3ndm$, where $n$ is the number of image tokens, $d$ is the hidden state size and $m$ is the intermediate size of the FFN. For latency, we report the average total inference time per sequence. For KV Cache memory usage, we report the average GPU memory consumption during inference. Table \ref{tab:eff} reports the results for each LVLM, averaged across eight benchmarks. CATP achieves the highest accuracy and shows excellent efficiency. It ranks second only to DivPrune in reducing FLOPs and KV Cache and delivers the largest cut in inference latency. As pointed out by \cite{pruning2}, latency is the most reliable indicator of token pruning efficiency because the execution cost of the method is not reflected in FLOPs or KV Cache. These results indicate that CATP brings real efficiency gains in multimodal ICL. Considering both its unique performance boost and its substantial efficiency gains, CATP stands out as an effective pruning method with strong practical potential for interleaved image-text tasks. 

\subsection{Ablation Study and Analysis}
\begin{figure}[t] 
  \centering                 
  \includegraphics[width=\columnwidth]{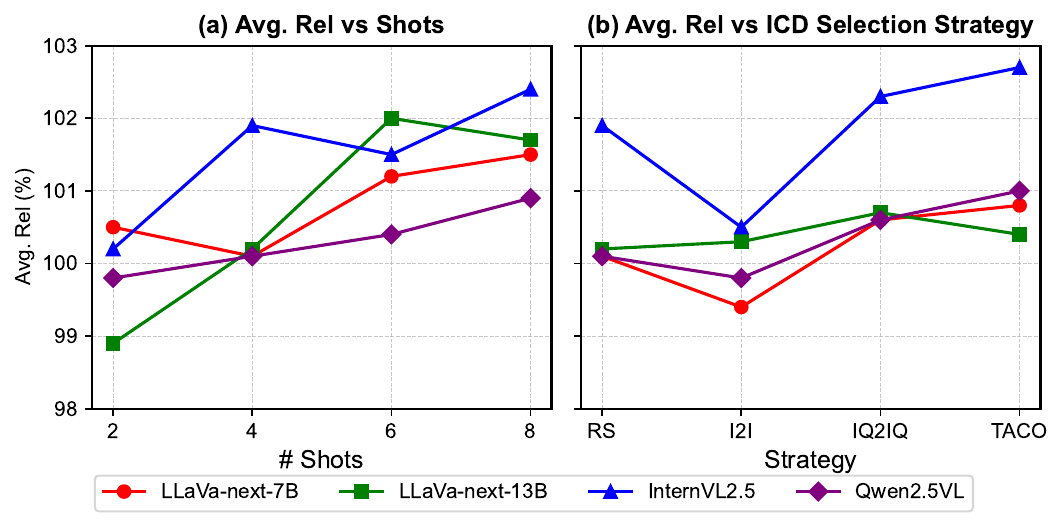}
  \caption{Performance of CATP on four LVLMs across diverse: (a) shot counts and (b) ICD selection strategies.}             
  \label{ab1}        
\end{figure}
All ablation studies are under a pruning ratio of $R=77.8\%$.
\begin{table}[t]
    \centering
    \small
    \setlength{\tabcolsep}{1.2pt}
    \scalebox{0.9}{
    \begin{tabular}{c c c c c}
        \toprule[1.5pt]
        \textbf{CATP} & \textbf{LLaVA-n-7B} & \textbf{LLaVA-n-13B} & \textbf{InternVL2.5} & \textbf{Qwen2.5VL} \\
        \hline
        Full & 100.1\%& 100.2\%& 101.9\%& 100.1\% \\
        w/o Stage 1  &96.7\%  & 95.4\% & 98.3\%& 98.0\% \\
        w/o Stage 2  & 98.2\%& 96.9\% & 98.9\% & 97.6\%\\
        \bottomrule
    \end{tabular}}
    \caption{Ablation results of the two stages in CATP.}
    \label{abla:stage}
\end{table}
\textbf{Adaptability to diverse in-context sequence configurations.} 
In practical multimodal ICL, both the number of ICD shots and the ICD selection strategy vary across applications. Since sequence construction strongly affects performance \cite{sen2,sen1}, we evaluate CATP under diverse configurations to assess its robustness. Figure \ref{ab1}(a) shows that CATP performs better as shot count increases, due to its ability to handle rising token redundancy and enhance reasoning by keeping only informative tokens. As LVLMs support longer contexts, CATP’s benefits can grow accordingly. Its inference speedup also scales nonlinearly with shot count (see Appendix 3.2). Figure \ref{ab1}(b) reports results under four ICD selection strategies. RS denotes the random sampling used in our main experiments. I2I and IT2IT are similarity-based methods commonly used in modern retrieval-augmented generation (RAG) systems, based on image-only and joint image-text similarity, respectively. TACO \cite{taco} is a SOTA ICD selection model. CATP brings greater performance gains as the quality of the sequence improves, suggesting that it can be integrated into existing systems to provide incremental benefits.

\textbf{Impact of each stage.} As shown in Table \ref{abla:stage}, across all LVLMs, removing either stage and letting the other perform the full pruning ratio $R$ leads to performance drops. This confirms that CATP’s two-stage adaptive pruning is essential. Meanwhile, removing Stage 1 results in a larger performance decrease than removing Stage 2. We attribute this to Stage 1 filtering out less informative image tokens before the decoder, allowing the attention and relevance computations in Stage 2 to better reflect token importance and avoid bias introduced by redundant tokens. In Appendix 3.3, we conduct ablation studies on the design details of each stage.

\begin{figure}[t] 
  \centering                 
  \includegraphics[width=\columnwidth]{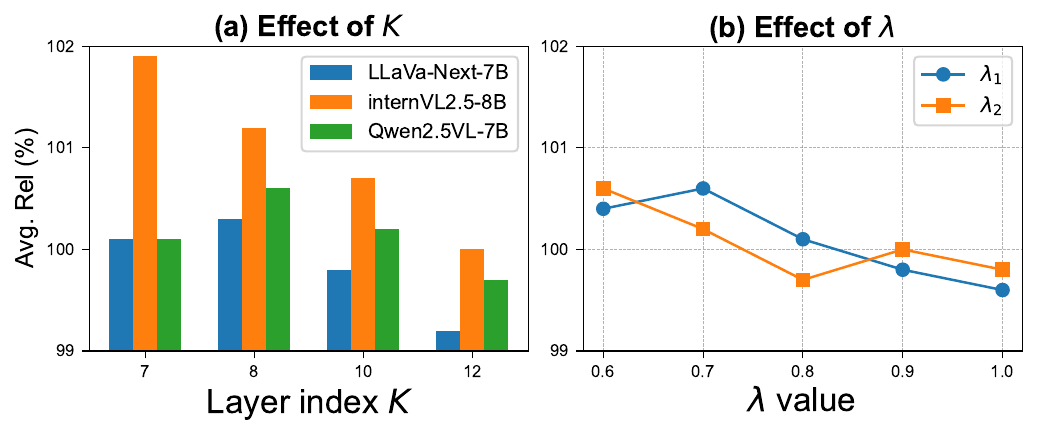}
  \caption{Average performance of CATP under different settings of $K$, $\lambda_1$, and $\lambda_2$.}      
  \label{hyper}        
\end{figure}

\textbf{Impact of hyperparameters.} Figure \ref{hyper} shows how CATP’s performance varies with different hyperparameter settings. Overall, CATP remains relatively stable, demonstrating robustness to hyperparameter choices. When $K$ is set to middle layers, performance drops, indicating that shallow layers contribute more to query-guided reasoning and that attention differences in these layers better reflect the importance of tokens. Further analysis of $\lambda$ is provided in Appendix 3.4.

\begin{figure}[t] 
  \centering                 
  \includegraphics[width=0.9\columnwidth]{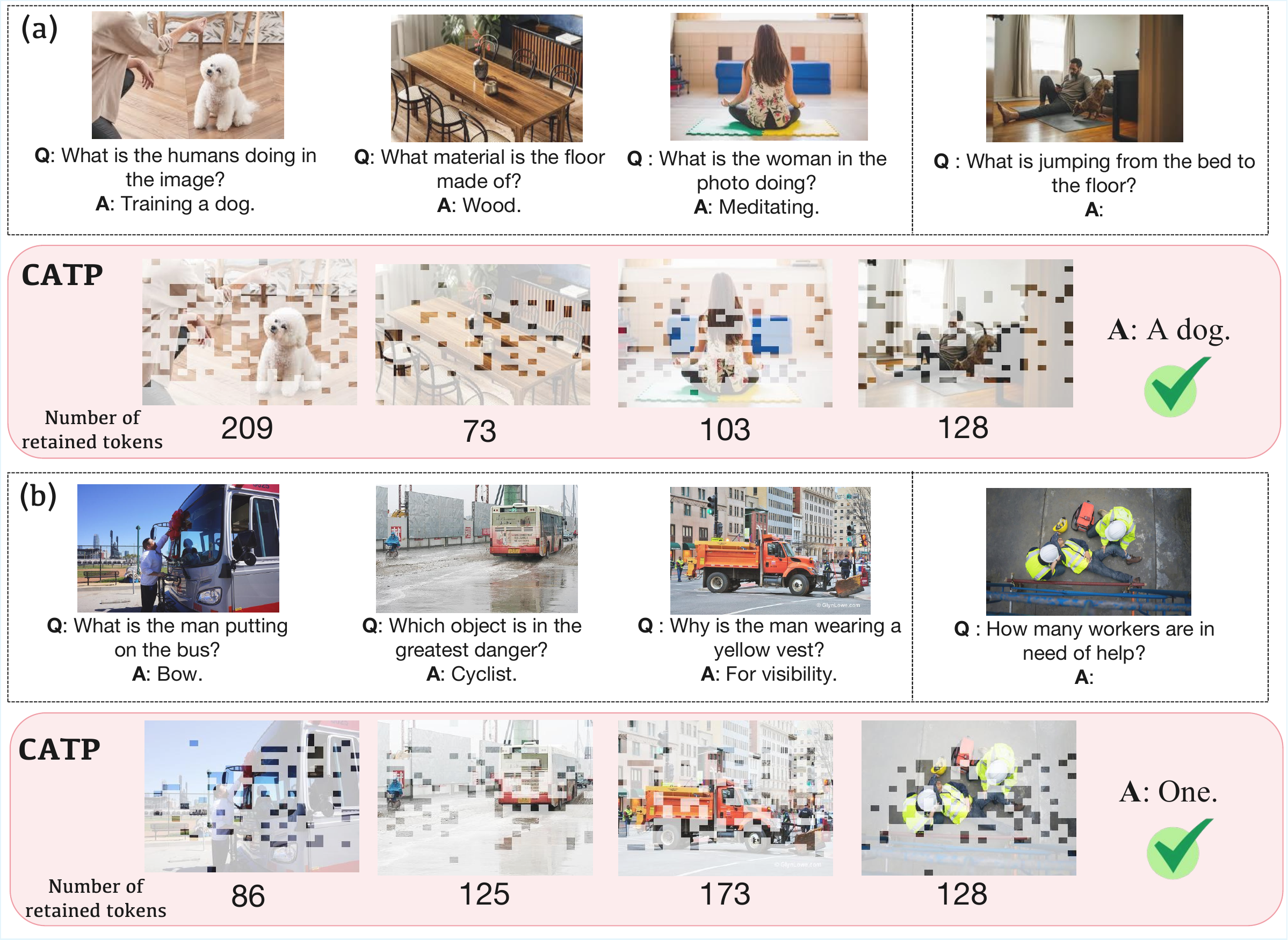}
  \caption{Qualitative visualizations of CATP on two cases.}           
  \label{quli}           
\end{figure}

\textbf{Qualitative visualizations.} The results in Figure \ref{quli} further demonstrate the ability of CATP to adapt seamlessly to multimodal ICL. Even after pruning a large number of image tokens, CATP still forms an effective sequence. Additional visualizations are provided in Appendix 3.5.
\section{Conclusion}
In this paper, we introduce Contextually Adaptive Token Pruning (CATP), a novel and training-free image token pruning method tailored to multimodal ICL. In the two-stage pruning process, CATP adaptively identifies the image tokens that are most important to the entire ICL process based on the input in-context sequence and discards the rest. This leads to improvements in both performance and efficiency. CATP addresses an important gap in existing work on multimodal ICL and offers a reliable and efficient solution to enhance the capability of LVLMs when handling complex inputs. We believe CATP provides a solid foundation and valuable insights for advancing LVLM capabilities.

\section*{Acknowledgments}
We acknowledge the computing resources provided by NSF ACCESS.
\bibliography{aaai2026}
\clearpage
\appendix
\begin{figure*}[ht] 
  \centering                 
  \includegraphics[width=\textwidth]{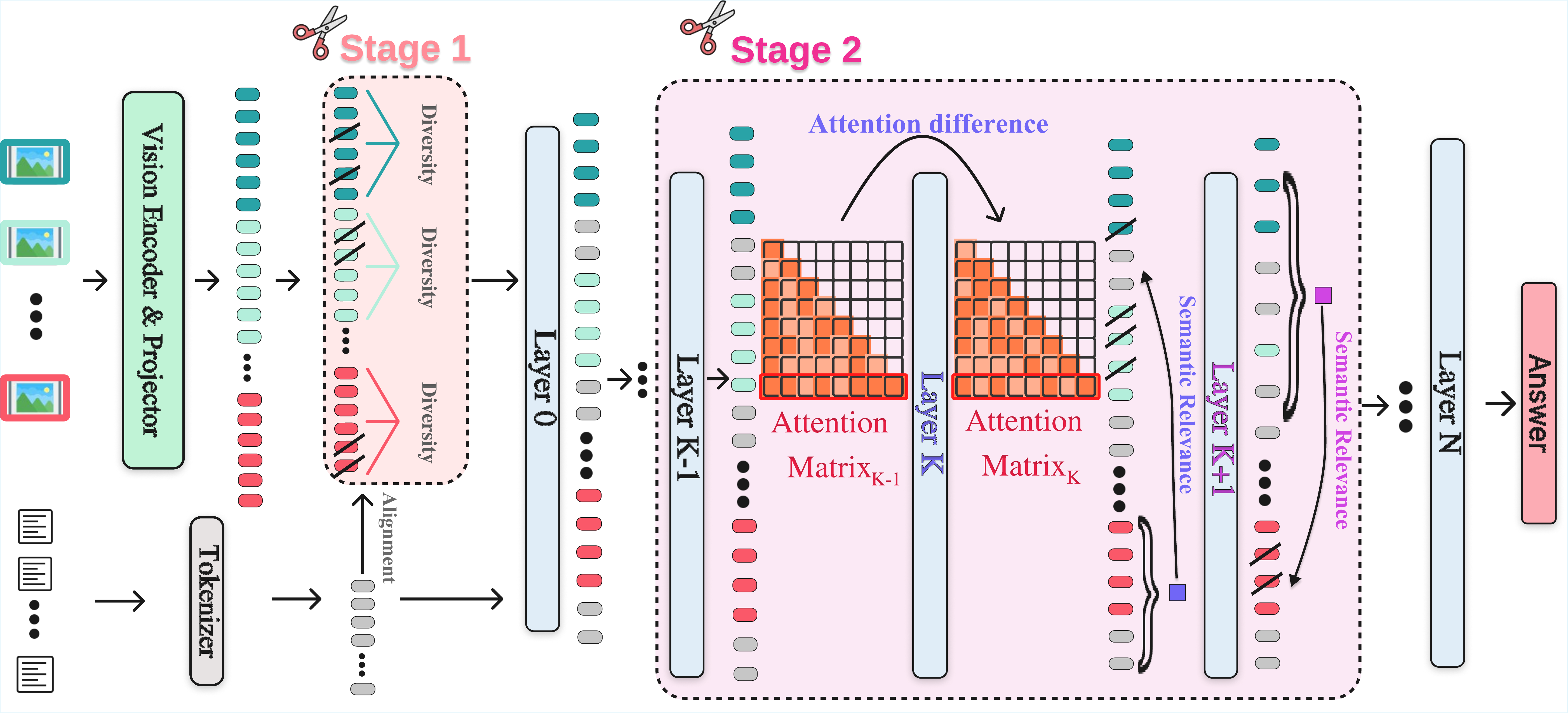}
  \caption{Overview pipeline of CATP. CATP consists of two stages. Stage 1 is applied between the projector and the decoder. It measures the importance of image tokens using feature distribution diversity and alignment with their paired text. Stage 2 operates in two shallow decoder layers. It first prunes the ICD image tokens by combining inter-layer attention differences with semantic relevance, and then prunes the query sample’s image tokens in the following layer.} 
  \label{main_d}           
\end{figure*}
\section{Appendix}
\subsection{1 $\quad$ Figure of Pipeline}
We present a detailed version of CATP's pipeline in Figure \ref{main_d}.
\subsection{2 $\quad$ Attention Shifts in Multimodal ICL}
\begin{figure}[ht] 
  \centering                 
  \includegraphics[width=\columnwidth]{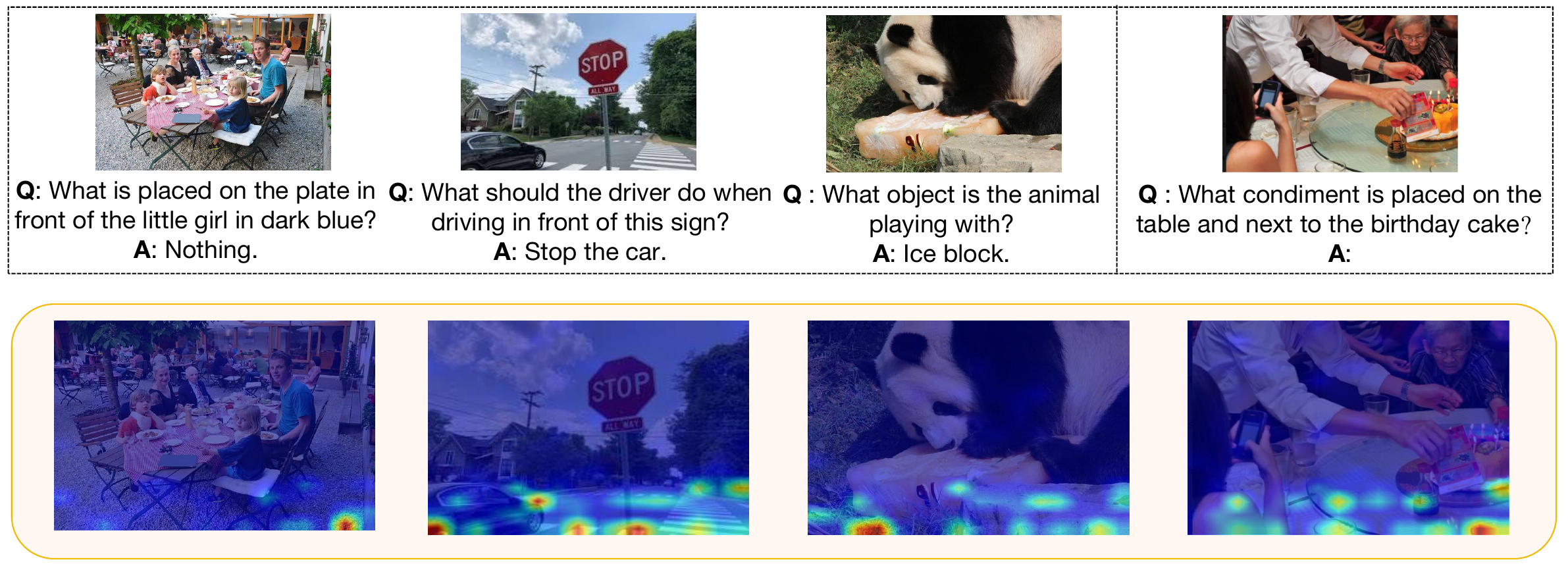}
  \caption{Due to attention shifts, LVLM may incorrectly concentrate its attention on image tokens located nearer the text, and this problem is amplified in multimodal ICL.} 
  \label{shift}        
\end{figure}

Attention shift \cite{as2,limi1} is the layer-wise drift of cross-modal attention within LVLM's decoder: visual tokens that initially draw little attention can gradually absorb a large fraction of the text-to-vision attention budget, so the amount of attention they receive no longer reflects their true saliency. Because transformers renormalize attention at every block, even small relocations compound, so by the final layers the pattern of “important” image tokens can be almost inverted relative to the shallow view. Empirically, this drift appears as clusters of redundant patches and tends to concentrate attention weights disproportionately on tokens that are closer to the text, whereas distinctive but more distant regions are overlooked, as illustrated in Figure \ref{shift}.

Thus, metrics that rely on raw attention inherit this instability \cite{ada,logic}. Average or cumulative attention scores, top-k frequency counts, and similar heuristics implicitly assume that a token’s weight is stationary. When that assumption fails, the metrics overestimate tokens that merely inherit migrated attention and underestimate those whose relevance surfaces late. Pruning or saliency mapping driven by such metrics either retains many near-duplicate patches, wasting computation, or excises genuinely informative ones, degrading vision-language performance. The mismatch between predicted and actual redundancy also grows with model depth.

The problem worsens in multimodal ICL, where images and text are interleaved. Relative-position or rotary encodings bias attention toward tokens that sit closest to the current textual span, so each newly inserted image block initially receives an artificial boost. As more text and images accumulate, attention repeatedly migrates toward the most adjacent visual tokens, pushing earlier images into a low-attention regime regardless of their semantic relevance. This positional bias amplifies attention shifts, making importance estimates increasingly local and myopic, and causing pruning methods based on attention to fail far more often in long, interleaved contexts than in single-image tasks.
\subsection{3 $\quad$ Experiments}
\subsubsection{\textbf{3.1 Setup Details}}

\paragraph{Benchmarks.} We evaluate CATP across eight benchmarks to comprehensively assess its effectiveness:
\begin{itemize}
    \item \textbf{VQAv2}: VQAv2 contains 443,757 samples in the training set and 214,354 in the validation set. It is a classic VQA benchmark that tests a model’s ability to understand both the image and the question across diverse real-world scenarios. The images are sourced from the MSCOCO dataset, and the evaluation metric is Accuracy.
    \item \textbf{VizWiz}: VizWiz contains 20,523 samples in the training set and 4,319 in the validation set. It presents a more challenging setting with lower-quality images, along with many unanswerable cases, pushing models to handle uncertainty based on the format learned from ICDs. Its evaluation metric is Accuracy.
    \item \textbf{OK-VQA}: OK-VQA contains 9,055 samples in the training set and 5,000 in the validation set. It requires the model to incorporate external knowledge beyond the image content and the context to generate correct answers. The evaluation metric is Accuracy.
    \item \textbf{GQA}: GQA contains 943k samples in the training set and 132k in the validation set. This benchmark requires the model to perform multi-step, compositional reasoning when answering VQA questions, rather than relying on answer priors. The evaluation metric is Accuracy.
    \item \textbf{HatefulMemes}: HatefulMemes contains 8,500 samples in the training set and 500 in the validation set. It is designed to reflect real-world challenges found in internet multimodal data. The task requires the model to jointly interpret both the image and the overlaid text to detect instances of hate speech. The evaluation metric is ROC-AUC.
    \item \textbf{MMStar}: MMStar consists of 1,500 hand-picked, vision-indispensable samples balanced across 6 core skills and 18 fine-grained axes. Every item was filtered from 22,000 candidates to eliminate data leakage and questions answerable by language alone, so models cannot “cheat” with memorized knowledge. Since it does not provide a predefined training and validation split, we divide the data using a 7:3 ratio. The evaluation metric is Accuracy, and the reported results are weighted averages across its six subtasks.
    \item \textbf{MME-Realworld}: We use the Perception subset of MME-Realworld, which contains 20,767 QA pairs spanning five task domains. This benchmark targets practical scenarios and offers the highest average image resolution among current VQA datasets. Because no official split is provided, we randomly divide the data into training and validation with a 7:3 ratio. The evaluation metric is Accuracy, and the reported scores are weighted averages over the five domains.
    \item \textbf{VL-ICL}: VL-ICL Bench is a comprehensive evaluation suite tailored for multimodal ICL, encompassing both image-to-text and text-to-image tasks. It tests a broad range of capabilities, from fine-grained perception and reasoning to fast concept binding, all using a few ICDs. In this work, we only employ the image-to-text split of VL-ICL Bench, which includes Fast Open MiniImageNet, CLEVR Count Induction, Operator Induction, TextOCR, Interleaved Operator Induction, and Matching MiniImageNet. 
\end{itemize} 

\paragraph{Models.} In this work, we evaluate CATP on four LVLMs with diverse configurations:
\begin{itemize}
    \item \textbf{LLaVA-Next-7B}: LLaVA-Next-7B uses CLIP ViT-L/336 as the vision encoder and Mistral-7B as the LLM backbone. It is instruction-tuned on high-quality visual-instruction pairs that emphasize OCR and logical reasoning while reusing the efficient LLaVA-1.5 training recipe. It converts each image into up to 2,880 tokens for high-resolution inputs. Considering its context length, we set the number of tokens per image to 576.
    \item \textbf{LLaVA-Next-13B}: LLaVA-Next-13B uses CLIP ViT-L/336 as its vision encoder and Vicuna-1.5-13B as the LLM backbone. Leveraging the same two-stage recipe as LLaVA-1.5, it is instruction-tuned on $\approx$ 1.3M curated visual-instruction pairs (558K connector warm-up samples plus 760K full fine-tuning samples) that focus on OCR, document/chart reasoning, and diverse real-user prompts, giving the model stronger factual and logical skills while remaining data-efficient. It can also convert each image into up to 2,880 tokens, and we set the number of tokens per image to 576.
    \item \textbf{InternVL2.5-8B}: InternVL2.5-8B uses InternViT-300M as the vision encoder and InternLM2.5-7B chat as the LLM backbone. It uses a three-stage curriculum (MLP warm-up → optional ViT incremental learning → full instruction tuning) combined with dynamic high-resolution tiling, JPEG-robust augmentation, and a stringent data-filtering pipeline for training. It can adjust the number of tokens per image based on input resolution. We set the token budget to the range from 576 to 1280. InternVL2.5 applies Grouped Query Attention (GQA), reducing the size of KV Cache significantly.
    \item \textbf{Qwen2.5VL-7B}: Qwen2.5VL-7B uses a window-attention ViT that supports dynamic spatial/temporal sampling and multi-resolution RoPE as the vision encoder and Qwen2.5-7B as the LLM backbone. It is pretrained on 4.1 T text tokens plus large-scale image + video data, then instruction-tuned for chat-style vision-language tasks. It can also adjust the number of tokens per image based on input resolution. We set the token budget to the range from 576 to 1280. Qwen2.5VL also applies GQA to reduce the size of KV Cache.
\end{itemize} 

\paragraph{Baselines.} As an important research topic, numerous image token pruning methods have been proposed, all showing effectiveness on single-image tasks. We select seven representative approaches that can be adapted to multimodal ICL as our baselines:
\begin{itemize}
    \item \textbf{FastV}: FastV shows that in the middle and deep layers of LVLMs, image tokens receive much less attention than text tokens, which indicates severe redundancy in these layers. Therefore, the method concentrates on early-stage token pruning by exploiting the attention that image tokens obtain from all other tokens. We apply FastV at the 2nd layer of the decoder.
    \item \textbf{HiRED}: HiRED first allocates a token budget to each image partition according to the attention received by the global image token, for example the CLS token. It then selects the most informative tokens within every partition, thereby achieving spatially aware token reduction.
    \item \textbf{FitPrune}: FitPrune reduces the visual token sequence inside the multi-head attention of every decoder layer. It decides which tokens to prune through binary search on attention statistics gathered from a calibration set, aiming to minimize the distributional gap between activations before and after pruning.
    \item \textbf{VTW}: Visual Tokens Withdrawal (VTW) is motivated by the observation that visual information is progressively transferred to text tokens in the middle and deep layers, leaving the remaining image tokens redundant. VTW therefore removes all image tokens after a designated layer, allowing only text tokens to participate in subsequent computation. We set this layer index to 10 for LLaVA-Next-7B, InternVL2.5-8B, and Qwen2.5VL-7B, and to 12 for LLaVA-Next-13B.
    \item \textbf{DivPrune}: DivPrune frames image token pruning after the projector as a Max-Min diversity problem, aiming to choose a subset of tokens that maximizes diversity among the selected tokens.
    \item \textbf{SparseVLM}: SparseVLM first identifies the key text tokens before the decoder. It then ranks token importance within the decoder through cross modal attention, applies adaptive sparsity ratios, and recycles the pruned tokens.
    \item \textbf{PLPHP}: PLPHP also addresses the multi-image scenario for visual token pruning. It adopts a two-level strategy comprising Layer Level Retention Rate Allocation and Head Level Vision Token Pruning. This design dynamically adjusts the token retention rate at each layer and allows different heads within the same layer to independently preserve critical context. We use a uniform hyperparameter setting of $(r,\Delta r,\alpha,\beta)=(0.4,0.3,0.2,0.15)$.
\end{itemize}

\subsubsection{\textbf{3.2 Efficiency with More Shots}}
\begin{table}[t]
    \centering
    \begin{tabular}{c|cccc}
        \toprule
        \textbf{Model} & \textbf{2-shot} & \textbf{4-shot} & \textbf{8-shot} & \textbf{16-shot} \\
        \midrule
        LLaVA-Next-7B      & 7.4\% & 16.5\% & 20.7\% & - \\
        LLaVA-Next-13B     & 4.1\%& 8.6\% & 18.3\%& - \\
        InternVL2.5-8B    & 4.5\% & 8.0\% & 14.9\% & 25.6\%\\
        Qwen2.5VL-7B     & 5.2\% & 10.0\% & 16.5\% & 28.0\% \\
        \bottomrule
    \end{tabular}
    \caption{Relative inference latency reductions (vs. Vanilla) for each model at shot counts of 2, 4, 8, and 16.}
    \label{app:effi}
\end{table}

As shown in Table \ref{app:effi}, the efficiency gains brought by CATP become more evident as the ICD shot count in the input sequence increases. With the growing demand for long-context tasks, the context length of current generation LVLMs has greatly expanded. This extension permits the inclusion of many multimodal ICDs, more fully capturing user needs and even enabling many-shot ICL, which is an important part of retrieval-augmented generation (RAG) systems. CATP can be effectively incorporated into these scenarios to mitigate inference latency, thereby promoting its adoption in a broader range of industrial-grade applications \cite{xiong1,xiong2}.

\subsubsection{\textbf{3.3 Additional Ablation Studies}}
\begin{table}[t]
    \centering
    \small
    \setlength{\tabcolsep}{1.2pt}
    \scalebox{0.9}{
    \begin{tabular}{c c c c c}
        \toprule[1.5pt]
        \textbf{CATP} & \textbf{LLaVA-n-7B} & \textbf{LLaVA-n-13B} & \textbf{InternVL2.5} & \textbf{Qwen2.5VL} \\
        \midrule
        Full & 100.1\%& 100.2\%& 101.9\%& 100.1\% \\
        w/o Stage 1 & 98.2\%& 96.9\% & 98.9\% & 97.6\%  \\
        w/o Stage 2  &96.7\%  & 95.4\% & 98.3\%& 98.0\% \\
        \midrule
        \multicolumn{5}{c}{Stage 1} \\
        w/o $\mathcal{F}_{\text{align}}(Y_i)$ & 97.2\%& 96.1\% & 98.1\% & 98.4\%\\
        w/o $\mathcal{F}_{\text{div}}(Y_i)$  & 98.3\%& 97.8\% & 99.5\% & 99.2\%\\
        \midrule
        \multicolumn{5}{c}{Stage 2} \\
        w/o $t_{ql}$  & 99.2\%& 98.7\% & 99.6\% & 99.1\%\\
        w/o $\Delta \mathcal{A}(c)$  & 98.1\%& 96.7\% & 98.8\% & 98.2\%\\
        w/o $\mathcal{S}_{\text{query}}(q)$  & 97.5\%& 97.2\% & 98.5\% & 99.0\%\\
        isolate ICDs & 97.4\%& 97.7\% & 98.0\% & 99.4\%\\
        \bottomrule
    \end{tabular}}
    \caption{Ablation results of the key designs in the two stages of CATP. 'w/o $\mathcal{F}_{\text{align}}(Y_i)$' and 'w/o $\mathcal{F}_{\text{div}}(Y_i)$' mean that Stage 1 keeps only one of the two scoring functions. 'w/o $t_{ql}$' means that Stage 2 computes attention with a random token from the query sample instead of its last token. 'w/o $\Delta\mathcal{A}(c)$' means that Stage 2 skips the attention difference and directly uses the attention from layer $K$. 'w/o $\mathcal{S}_{\text{query}}(q)$' means that Stage 2 prunes the query sample image tokens together with the ICD tokens at layer $K$ instead of pruning them separately at layer $K+1$. ‘Isolate ICD’ means pruning the image tokens of each ICD individually with a fixed ratio.} 
    \label{app:abla}
\end{table}

\begin{figure}[t] 
  \centering                 
  \includegraphics[width=0.95\columnwidth]{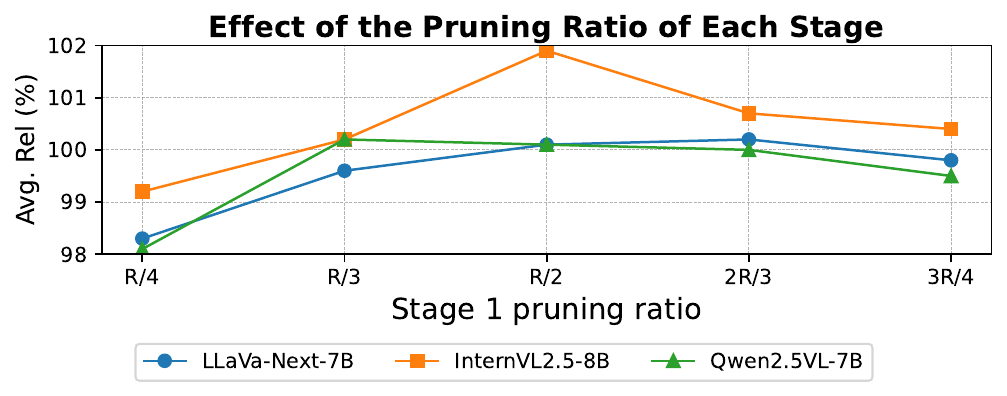}
  \caption{Average performance of CATP as the pruning ratios of the two stages vary.}      
  \label{hyper3}        
\end{figure}

From Table \ref{app:abla} we draw the following conclusions about the key designs of CATP: (1) In Stage I, jointly considering alignment and diversity is essential. In a multimodal ICL setting, diversity is slightly more important. This indicates that visual information can influence overall ICL directly without always coupling with text, which further highlights the need to refine visual features through image token pruning.
(2) The last token of the query sample already represents the full semantics of the query in the middle layers, so the attention received by this single token can serve as an indicator of each image token’s contribution to answer generation.
(3) Consistent with our analysis of existing attention-based methods, static attention from a single layer is insufficient for measuring token importance in multimodal ICL; using attention difference better reflects the interactions that arise during multimodal ICL.
(4) Pruning the query sample and the ICDs separately captures these interactions more effectively and thus improves CATP performance.
(5) Since different ICDs contribute unequally to ICL, sharing a common pruning ratio across their image tokens as a whole is necessary, which enables more fine-grained token pruning tailored to multimodal ICL. Furthermore, we investigate the influence of the pruning ratio allocation in each stage on CATP performance, as illustrated in Figure \ref{hyper3}. We find that the impact is model-specific, and allocating the pruning ratio of $R/2$ to each stage yields the best general performance. This result demonstrates the built-in balance between the two stages in the design of CATP.

\subsubsection{\textbf{3.4 Analysis of $\lambda_1$ and $\lambda_2$}}
\begin{figure}[t] 
  \centering                 
  \includegraphics[width=0.95\columnwidth]{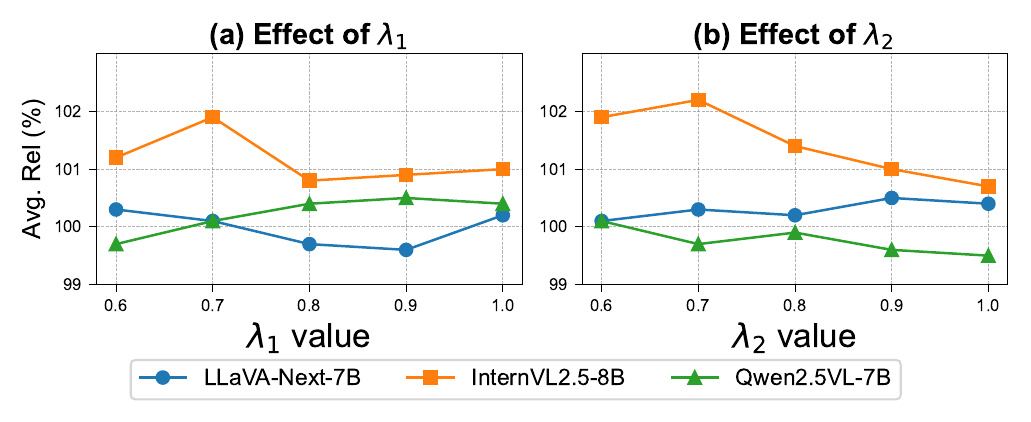}
  \caption{Average performance of CATP under different settings of $\lambda_1$ and $\lambda_2$.}      
  \label{hyper2}        
\end{figure}

When $\lambda_1$ increases, Stage 1 of CATP places greater emphasis on image-text alignment. When $\lambda_2$ increases, Stage 2 prunes the image tokens in the ICDs with heightened focus on semantic relevance. Figure \ref{hyper2} provides a closer look at how the two balancing coefficients affect different LVLMs. Panel (a) shows that a moderate emphasis on image-text alignment ($\lambda_1 = 0.7$) yields the largest relative gain of about two percentage points for InternVL2.5-8B, whereas LLaVA-Next-7B prefers either a lower or the maximal setting and Qwen2.5VL-7B benefits most when $\lambda_1$ approaches~0.9.
Panel (b) indicates that giving more weight to semantic relevance in Stage~2 brings clear improvements only up to a point.
InternVL2.5-8B again peaks at $\lambda_2 = 0.7$; further increases gradually erode its advantage.
LLaVA-Next-7B reaches its best value near $\lambda_2 = 0.9$, while Qwen2.5VL-7B performs best at the smallest $\lambda_2$ and declines steadily afterwards.
These trends confirm that the relative importance of alignment and relevance signals is model‐specific. 
\subsubsection{\textbf{3.5 Additional Qualitative Visualizations}}
\begin{figure}[ht] 
  \centering                 
  \includegraphics[width=\columnwidth]{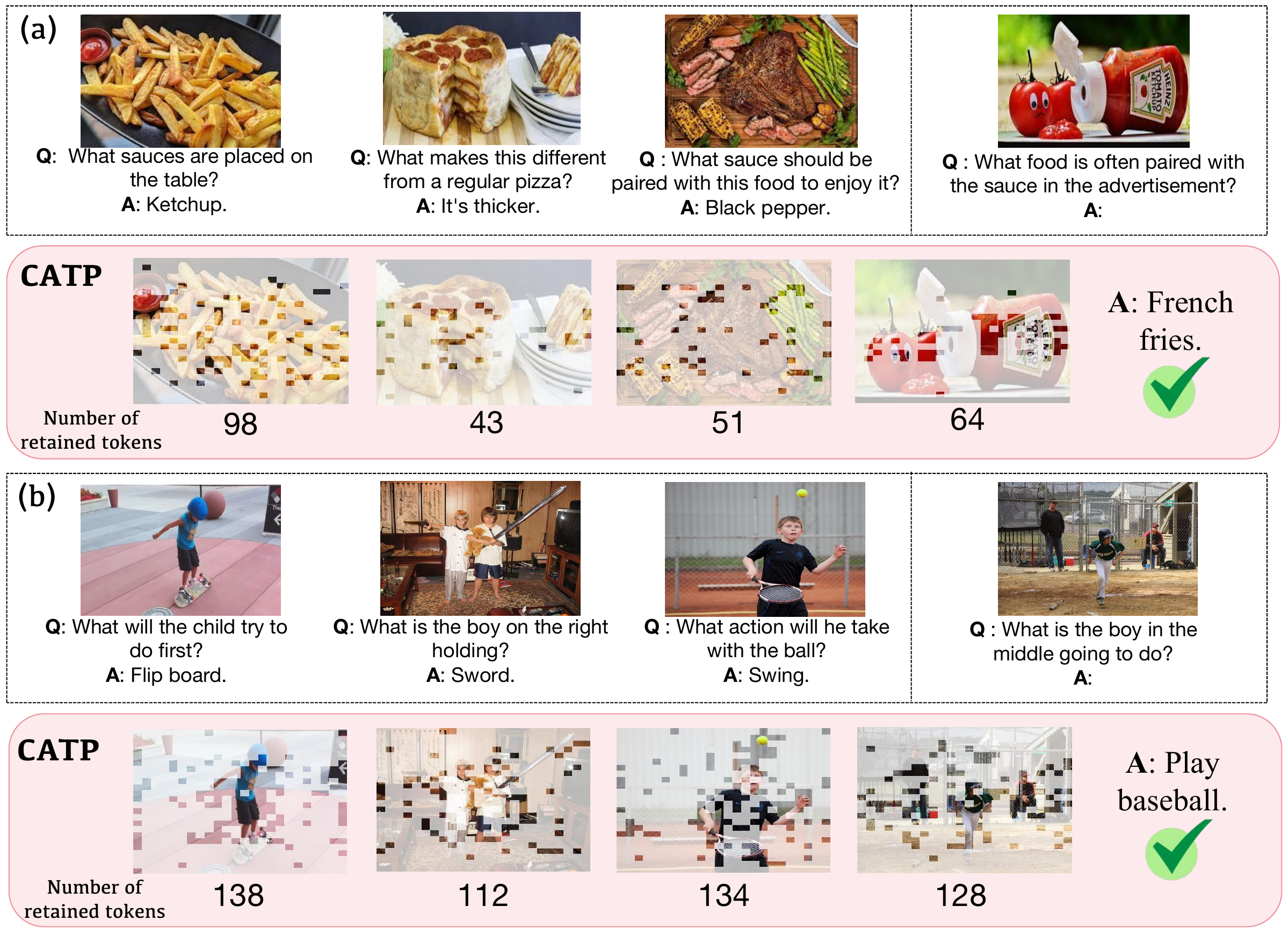}
  \caption{Qualitative visualizations of CATP on two cases.} 
  \label{quali2}        
\end{figure}
Figure \ref{quali2} presents additional qualitative visualizations that illustrate how CATP adapts to the contexts. Rather than pruning image tokens only according to the semantics inside each image-text pair, CATP evaluates their importance across the entire sequence. Thus, the tokens retained after pruning contribute most to the complete ICL process and provide accurate support for answering the query sample. For example, in the first case, CATP keeps the tokens linked to both “Ketchup” and “French fries” in the first image, making it the ICD with the greatest number of retained tokens. In the second case, the first image preserves more tokens tied to locating the person, while the third image keeps more tokens related to action recognition, and these selections jointly support the answer to the query sample.
\end{document}